\documentclass[conference]{IEEEtran}
\ifCLASSINFOpdf
\else
\fi
\usepackage{listings}
\usepackage{xcolor}
\usepackage{multirow}
\usepackage{array}
\usepackage{url}

\definecolor{codegreen}{rgb}{0,0.6,0}
\definecolor{codegray}{rgb}{0.5,0.5,0.5}
\definecolor{codepurple}{rgb}{0.58,0,0.82}
\definecolor{backcolour}{rgb}{0.95,0.95,0.92}

\lstdefinestyle{mystyle}{
    backgroundcolor=\color{backcolour},   
    commentstyle=\color{codegreen},
    keywordstyle=\color{magenta},
    numberstyle=\tiny\color{codegray},
    stringstyle=\color{codepurple},
    basicstyle=\ttfamily\footnotesize,
    breakatwhitespace=false,         
    breaklines=true,                 
    captionpos=b,                    
    keepspaces=true,                 
    numbers=left,                    
    numbersep=5pt,                  
    showspaces=false,                
    showstringspaces=false,
    showtabs=false,                  
    tabsize=2
}

\usepackage{graphicx}
\usepackage{xcolor, color, colortbl}
\usepackage{tabularx}
\usepackage{tikz}
\newcommand*\circled[1]{\tikz[baseline=(char.base)]{
            \node[shape=circle,fill,inner sep=0.4pt] (char) {\textcolor{white}{#1}};}}

\definecolor{royalpurple}{RGB}{120, 81, 169}
\newcommand{\AT}[1]{\textcolor{blue}{#1}}
\newcommand{\notes}[1]{\textcolor{royalpurple}{#1}}
\newcommand{\mycomment}[1]{}

\begin{document}
%
\title{Accelerated development and deployment of AI inference services on the heterogeneous edge-cloud continuum with TF2AIF}

\title{TF2AIF: Facilitating development and deployment of accelerated AI models on the cloud-edge continuum
}


\author{\IEEEauthorblockN{Aimilios Leftheriotis$^{\star}$, Achilleas Tzenetopoulos$^{\dagger}$, George Lentaris$^{\dagger}$$^{\ddagger}$, Dimitrios Soudris$^{\dagger}$, Georgios Theodoridis$^{\star}$}
\IEEEauthorblockA{
$^{\star}$\textit{University of Patras, Greece} \\
$^{\dagger}$\textit{National Technical University of Athens, Greece} \\
$^{\ddagger}$\textit{University of West Attica, Greece}
}

}

\maketitle

\begin{abstract}
    
The B5G/6G evolution relies on connect-compute technologies 
and highly heterogeneous clusters with HW accelerators,
which require specialized 
coding to be efficiently utilized.
The current paper proposes a custom tool for generating
multiple SW versions of a certain AI function
input in high-level language, e.g., Python TensorFlow,
while targeting multiple diverse HW+SW platforms.
TF2AIF builds upon disparate tool-flows to
create a plethora of relative containers 
and enable the system orchestrator to
deploy the requested function on any peculiar node in
the cloud-edge continuum,
i.e., to leverage the performance/energy
benefits of the underlying HW upon any circumstances.
TF2AIF fills an identified gap in today's ecosystem
and facilitates research on resource management or
automated operations, 
by demanding minimal time or expertise from users. 

\end{abstract}


%
\IEEEpeerreviewmaketitle

\section{Introduction}
\footnote{© 2024 IEEE.  Personal use of this material is permitted.  Permission from IEEE must be obtained for all other uses, in any current or future media, including reprinting/republishing this material for advertising or promotional purposes, creating new collective works, for resale or redistribution to servers or lists, or reuse of any copyrighted component of this work in other works. Accepted by EUCNC \& 6G Summit 2024.} The advent of B5G/6G represents a spectacular 
shift in mobile communication technologies 
with advanced networks promising enhanced performance in terms of data rates, latency, and device connectivity.
To achieve these goals,
the B5G/6G world will rely on the convergence between connect and compute resources of growing heterogeneity,
ranging from low-power to high-performance processors, 
server to embedded HW accelerators,
open- to closed-source SW tools,
and furthermore, 
spanning from cloud- to near- to the far-edges of the network \cite{kartsakli2023ai}.
Considering the diversity of infrastructures, 
the B5G/6G integration should ensure that SW services are efficiently deployed to the most suitable HW across the cloud-edge continuum and managed seamlessly to
sustain QoS, 
all while balancing user demands and HW constraints.





One of the key functionalities being deployed on B5G/6G is Artificial Intelligence (AI),
either for operational decisions such as network slicing \cite{blanco2023ai}
or for client applications such as 
object detection on cameras \cite{patterson2022hivemind}.
Allocating and managing computationally-intensive AI models across numerous heterogeneous nodes 
is a great challenge,
which becomes even more complicated when
considering the varying optimization level 
or even the code unavailability
of a single AI Function (AIF)
for multiple diverse HW-SW platforms.
That is to say,
the high-level program that an AIF is usually provided in,
e.g., Python TensorFlow (TF),
is neither ready nor optimal for execution
on diverse accelerators or processor ISA's;
furthermore,
the AIF's algorithmic nature itself is not equally amenable
to efficient execution by every type of underlying HW.
However, 
heterogeneous code availability and optimization 
are prerequisites for sufficient performance/energy in B5G/6G, 
and thus, tools that automate/facilitate
us to generate appropriate multi-versioned  
executable AIFs for a plethora of 
platforms become critical.

In general, the complexities of resource allocation \& management 
are gradually alleviated from operators via
sophisticated backend systems offering Zero-touch deployment \cite{psaromanolakis2023mlops} and automatic orchestration.
Cloud-native technologies, such as Kubernetes \cite{burns2016borg}, are at the forefront of 
technology for
service orchestration 
in a flexible and uniform fashion \cite{basaras2023experimentally,tonini2023service}. 
State-of-the-art research 
explores orchestration 
strategies,
alas, with limitations in heterogeneity:
e.g., 
\cite{ferikoglou2023iris} primarily addresses homogeneous VMs
and \cite{giagkos2023darly} assumes variation only in CPU generations.
To additionally leverage HW acceleration,
we need a proliferation of 
multi-versioned AIFs,
as mentioned above.
However,
with a notable lack of such proposed automated tools,
integrating HW acceleration into diverse environments 
demands considerable expertise and time investment.

In the context of H2020 project AI@EDGE \cite{9482440},
among others, 
a connect-compute platform was developed
to support AI over heterogeneous compute clusters at
the far- and near-edges of B5G. 
The platform was concerned 
with serverless computing,
cross-layer multi-connectivity,
AIF auto-configuration,
non-RT RIC,
security,
multi-tier orchestration,
and HW acceleration.
Specifically for the last two,
both at research and deployment phases,
the aforementioned automation gap 
underscored a need to design a custom tool
for facilitating development, deployment, 
and utilization of AIFs in diverse environments
offering a simplified procedure to end-users.

Accordingly,
the current paper presents the design and
evaluation of TF2AIF.
The proposed tool inputs 
initial Python TensorFlow functions to seamlessly generate 
ready-to-use 
AIF versions for multiple distinct platforms \& vendors: 
x86 CPU, ARMv8 CPU, server-class FPGA, high-end GPU, 
mobile GPU, embedded SoC FPGA
all while assuming a range of supported OS
and deployment scenarios. 
This set of SW+HW components was selected as representative
of today's landscape and is extensible via TF2AIF. 
Our AIF integration is automated through
custom containerization techniques,
diverse tool-flow chaining, 
meticulous driver/library combinations,
parameterized configurations,  
and extensive scripting. 
As a result, 
we derive dozens of accelerated AIFs
in a significantly decreased amount of development time,
e.g., 
20 AIFs in approx. 10min, 
with minimal user knowledge of vendor SW.
Such an output repository  enables 
design space exploration,
research of resource management algorithms,
and a multitude of orchestration decisions 
during the platform's operational phase.

The remainder of this paper includes a succint related work (Sec. II),
comprehensive motivation (Sec. III),
all design details of TF2AIF (Sec. IV), its evaluation (Sec. V), 
and conclusions.


\section{Related Work}
\label{sec:related}

In the heterogeneous cloud-edge continuum, 
key focus is given to the efficient management of AI functions,
which are essential both for network automation 
and user-centric applications.
In general, these AIF's necessitate specialized hardware for AI acceleration 
with capabilities for training, updating, and deploying Machine Learning models \cite{9482440}. 
The literature includes relevant works 
with a significant strand of research leveraging different model variants to cater to specific AI services  \cite{ferikoglou2023iris, giagkos2023darly, hudson2021qos, romero2021infaas}. 
Indicatively, \cite{ferikoglou2023iris, hudson2021qos} demonstrate the use of distinct model implementations for identical AIF functionalities, such as image classification, to meet varying accuracy and latency requirements.
Conversely, INFaaS \cite{romero2021infaas} is more specialized, targeting proprietary infrastructure and offering only a limited selection of platforms and model variants. 
Utilizing both near- and far-edge GPUs, \cite{basaras2023experimentally} showcases the deployment of an object detection AI inference service in a 5G context.
Meanwhile, MDPM \cite{wang2022mpdm} introduces the concept of co-existing model variants, but is constrained only to CPU and GPU configurations.

\mycomment{
\AT{INFAAS - model variant\cite{romero2021infaas}
    Triton: we provide an interface that makes pre- and post-processing very easy.}

\AT{RL on CPU \cite{asenov2023learning,giagkos2023darly}}
}

The landscape of tools supporting AI inference serving is rich  \cite{nvidia-triton, AWS, BentoML, TorchServe, Tf-serving}, with several solutions addressing different aspects of deployment and operational challenges. 
Nvidia Triton \cite{nvidia-triton} is a pivotal tool that facilitates the deployment of AI models across various 
GPU platforms, emphasizing flexibility and efficiency. 
It supports multiple frameworks as input and backend, including TensorFlow, PyTorch, and ONNX, catering to a wide range of use cases. 
Despite its versatility, Triton requires users to have substantial knowledge about certain frameworks, especially for leveraging acceleration features and implementing model quantization.
Additionally,  users are required to construct complex machine learning pipelines to implement pre/post-processing tasks. 
Finally, Triton's current support for edge devices is limited.
\mycomment{Contrastingly, TF2AIF expands the accessibility and ease of deploying AI models by:

Supporting acceleration frameworks on all platforms, not limited to GPU/AGX, thereby broadening the scope of hardware utilization.
Eliminating the need for framework-specific knowledge to use acceleration features, making it more accessible to users without deep technical expertise in specific AI frameworks.
Offering simple quantization support (INT8) that does not require framework-specific knowledge, facilitating efficient model deployment on resource-constrained devices.
Providing actual edge support through the use of TensorFlow Lite, optimized for edge devices with quantization, to achieve reduced memory usage and enhanced performance.
Simplifying the implementation of preprocess and postprocess tasks on the server, making TF2AIF a more comprehensive solution for AI serving needs.}
INFaaS Model Generator \cite{romero2021infaas} is an academic tool that allows users to provide a model, from which containers are created with different model variants (configurations), making the container implementation process transparent to the user. 
However, currently, it includes limited support for diverse hardware platforms (esp. edge devices) and lacks integrated quantization support, 
which is essential for optimizing models 
on resource-constrained platforms.
Moreover, INFaaS's approach to pre/post-processing tasks is not explicitly detailed, potentially complicating deployments requiring comprehensive data handling.

\section{Motivation}
During literature/market survey and 
 work in AI@EDGE, 
as mentioned in the introduction,
we identified a considerable void in the SW 
tools' ecosystem regarding 
the creation of multiple AIF versions for distinct underlying HW+SW platforms. 
The motivation to build a unified tool for B5G/6G is threefold.

\textbf{Increasing complexity:}
B5G/6G requires seamless deployment of AI models across diverse platforms, spanning from the Cloud to the far-edge. 
The complexities of platform architectures, framework libraries, APIs, and container environments present considerable challenges in leveraging heterogeneous clusters for AI applications.
Besides focusing only on designing ML services, 
engineers need to also get familiar with vendor-specific tools, e.g., Vitis AI, for the development and deployment of their application in order to fully exploit the underlying hardware.
Following the paradigm of container orchestrators,
which added a level of abstraction for seamless infrastructure provisioning and management, the B5G/6G community would benefit
from a tool alleviating users from the aforementioned complexities to develop and deploy AIFs. 
Hence, \textit{Objective \#1 is to democratize AI model deployment on heterogeneous devices.}

\textbf{Challenges for benchmarking limiting design space exploration:}
Due to the plethora of AI frameworks and hardware platform combinations, even experienced engineers need to allocate a considerable amount of time to get familiar with, develop, test, and validate AIFs.
This is a non-negligible barrier for conducting design space exploration or benchmarking, to verify the model-variant that fits best to the desired use case.
A tool that hides implementation details and provides safeguards, will assist developers to avoid common pitfalls, minimize debugging time,
and provide comprehensive benchmarking solutions in a 
much shorter time.
Therefore, \textit{Objective \#2 is accelerating prototyping, testing, and benchmarking with a vendor-neutral tool}.

\textbf{Limited support regarding inference serving scheduling systems:} 
The scheduling of AI inference serving in heterogeneous cluster environments is a burgeoning area of research. 
In the majority of cases, it is accomplished by utilizing frameworks like Nvidia Triton as backbones for server container creation \cite{cui2022dvabatch, Yang2021DeepRT}.
However, these are often limited in terms of hardware support and require substantial AI-framework expertise from developers, hindering the full potential of these scheduling systems. 
By broadening the range of model variants and extending their applicability across more hardware platforms, 
these orchestration solutions can significantly improve. 
Furthermore, 
the use of AI in the scheduling procedures of inference serving systems is limited, with current solutions 
relying mostly on empirical approaches
\cite{romero2021infaas, cui2022dvabatch, Yang2021DeepRT}. 
A tool facilitating the creation of a  
necessary variety and volume of data 
is essential for effective ML training 
also regarding these scheduling procedures. 
Thus, \textit{Objective \#3 is to improve and expand inference serving system on heterogeneous clusters}.
Finally, \textit{Objective \#4: is to enable AI-driven inference serving scheduling systems on the cloud-edge continuum}.

\mycomment{
Challenges:
\begin{itemize}
    \item seamless deployment of AI models across diverse platforms (Ch.1)
    \item efficient AI inference serving containers on edge devices (technology democratization) (Ch.1)
    \item Accelerate the workflow of developers (Ch.2)
    \item Rapidly test, evaluate, and benchmark multiple models across different platform (Ch.2)
    \item Facilitate Design space exploration (Ch.2)
    \item Backbone for inference serving systems (both current and create new ones on extra-heterogeneous systems) (Ch.3)
    \item Empowers research into machine learning-based scheduling systems for AI inference serving on heterogeneous platforms addressing the challenges of data scarcity and quality (Ch.4)
\end{itemize}
CH.1 = A
Ch.2 = B + C.B + C.C
Ch.3 = C.D + C.E
Ch.4 = F
}


\section{TF2AIF}
\label{sec:implementation}

\subsection{Overview}


To overcome the challenges and meet the objectives mentioned above, we introduce TF2AIF, a tool that facilitates the development and deployment of AIFs in the heterogeneous, HW acceleration-enabled cloud-edge continuum.
Our vendor-neutral, open-source\footnote{the code will be provided in case of acceptance of the current paper} tool to assist towards AI democratization, aims to provide a layer of abstraction for users to automatically generate their code without delving into complex details.
To achieve this, TF2AIF natively supports a variety of Accelerated Inference frameworks (AI-frameworks), such as TensorRT, as well as devices with several flavors of heterogeneity.
These include both heterogeneous CPU architectures (x86, ARM) and hardware accelerators (GPUs, FPGAs).
However, TF2AIF is designed with extensibility in mind, allowing easy accommodation of additional AI-frameworks and platforms. 
TF2AIF comprises about 13.6k lines of code, 7k of which are in Python, and the rest are scripts.

\begin{figure}[t]
	\centering
    \centering
    \includegraphics[width=.95\columnwidth]{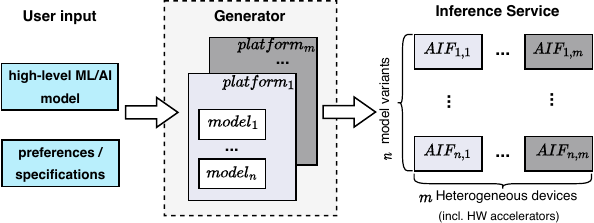}
     \caption{Automated model-variant generation}
     \label{fig:high-level-schema}
\end{figure}

In a high-level overview, TF2AIF, following the specifications demonstrated in Fig. \ref{fig:high-level-schema}, accepts a user's high-level Machine Learning (ML) model along with various user preferences. 
It then efficiently transforms these inputs into a set of model variants. 
Each variant is encapsulated with its corresponding environment in the form of container images.
TF2AIF also provides the respective client container for each AIF, as well as an integrated metrics collector that provides performance statistics.
Except for the model conversion, TF2AIF simplifies the complex processes of quantization, data pre/post-processing, and server-client communication.
Details on its components and features are discussed in Sections IV-C and IV-D.

\subsection{Supported Platforms \& Accelerated Inference Frameworks}

Since our target is extremely heterogeneous clusters that contain both server and edge devices, we have developed a wide range of different platforms. 
However, TF2AIF is designed to be easily extendable to any kind of platform device. 
Therefore, TF2AIF has been developed but is not limited to the following platform categories.
Common setups in the near-edge include physical servers equipped with x86 processors.
Additionally, hardware accelerators are connected via PCIE inside the server chassis.
The tool currently supports x86 CPU, data center FPGAs, and data center GPUs.
On the far edge, there are mostly used power-constrained devices.
These include embedded devices, usually with an ARM processor, because of their energy efficiency and cost-effectiveness.
Lately, devices with embedded hardware accelerators have emerged.
TF2AIF currently supports the NVIDIA Jetson Xavier AGX.

\begin{table}[t]
    \centering
    \caption{Inference Acceleration Frameworks by Platform and Precision}
    \begin{tabular}{llll}
        \hline
        \textbf{Name} & \textbf{Platform}                     & \textbf{Inf. Accel. Framework}                                & \textbf{Precision} \\ \hline
        AGX           & Edge GPU                              & ONNX w/ TensorRT\cite{ONNXwTRT}                                        & INT8               \\
        ARM           & ARM                                   & Tensorflow Lite\cite{TFlite}                                                           & INT8               \\
        CPU           & x86 CPU                               & Tensorflow Lite\cite{TFlite}                                                           & FP32               \\
        ALVEO         & Cloud FPGA                            & Vitis AI\cite{Vitis-AI}                                                                  & INT8               \\
        GPU  & GPU                                   & ONNX w/ TensorRT\cite{ONNXwTRT}                                       & FP32/FP16/INT8     \\ \hline
    \end{tabular}
    \label{table:inference_acceleration_precision}
\end{table}

Table \ref{table:inference_acceleration_precision} depicts the AI-framework-platform combinations that TF2AIF currently supports.
The objective is to combine a variety of different cloud and edge platforms, while simultaneously maximizing the performance gains by employing the appropriate AI-framework for each platform.
The selection of the specific AI-frameworks was guided by a benchmarking process on the AI-framework ecosystem, which indicated the most potent ones both from a performance and a functionality scope.
We associate a name to each AI-framework-platform combination. 
\mycomment{\textcolor{gray}{To accommodate the widespread use of GPUs and optimize performance, we implement two GPU AI-framework-platform combinations. 
While TF-TRT offers high performance but necessitates direct GPU connection, our alternative solution, though potentially less performant, removes this constraint, ensuring compatibility across all GPUs.}}

\mycomment{The reason behind the existence of two different combinations for GPUs is that the GPU platform is one of the most widely used and therefore choosing the most performant inference acceleration is important. 
However, the TF-TRT framework requires knowledge of the target GPU which requires the target GPU to be connected to the machine that TF2AIF is running on. 
Thus, we provide another solution which may be less performant, but alleviates this restriction and provides a container that is appropriate to run on every GPU.}

\subsection{Detailed Description of TF2AIF}
\label{sec:details}
\mycomment{
\begin{itemize}
    \item \AT{estiase pio polu sta components 1 pros 1.
input/output, pws to petyxainei, akolou8wntas th roh.}
\item \AT{apo to sxhma den fainetai to server/client final output}
\item \AT{to sxhma kalo 8a htan na ksexwrizei me ena tetragwno poio einai to ergaleio (ta aspra koutia)}
\end{itemize}
}
\begin{figure}[!t]
    \centering
    \includegraphics[width=0.99\columnwidth]{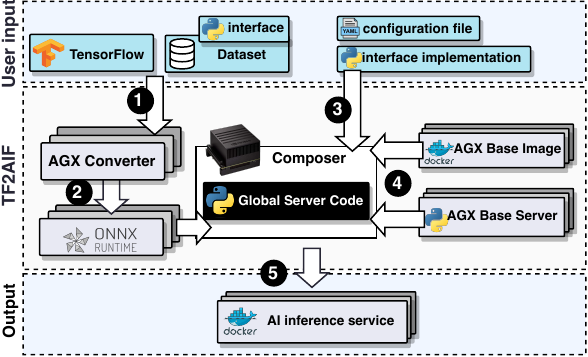}
    \caption{Example of AGX implementation}
    \label{fig_tf2aif_agx_single_example}
\end{figure}

Figure 2 illustrates the structure of TF2AIF, using, as an example, the generation of a server container for a far-edge node platform, i.e., NVIDIA Jetson AGX.
Specifically, the user provides inputs to the converter (\circled{1}), 
which produces the required model (\circled{2}).
Finally, the Composer, using the user-provided configuration files (\circled{3}), and some additional boilerplate code (\circled{4}) generates the AI inference service (\circled{5}).
The implementation process for each AI-framework-platform combination is the same as the example shown in Figure \ref{fig_tf2aif_agx_single_example}, with the appropriate changes in the tool resources (CPU converter for the CPU case etc.). 
TF2AIF implements every AI-framework-platform combination in parallel and reuses the same user inputs. 
The components, as well as the features of the tool, are thoroughly described in the rest of this subsection.

\mycomment{Figure \ref{fig_tf2aif_agx_single_example} depicts the implementation diagram for the AGX AIF container case, one of the six framework-platform combinations. 
Blue boxes represent user-provided inputs, the resources that are inherently part of the tool are depicted in white, and the output, which is the AIF container, is shown in beige. 
The implementation process for each framework-platform combination is the same as the AGX example shown in figure \ref{fig_tf2aif_agx_single_example}, with the appropriate changes in the tool resources \notes{(maybe change the color of the resources that change per platform?)} (ALVEO converter for the ALVEO case etc.). 
TF2AIF implements every framework-platform combination in parallel and reuses the same user inputs.}

\textbf{Converter:}
The Converter is a platform-specific component that is responsible for the conversion of the given TensorFlow model to the appropriate format (e.g., ONNX runtime).
Additionally, AI-frameworks facilitate or necessitate model quantization, a process where the model's weights are converted from the standard FP32 format to either FP16 or INT8. 
This transformation optimizes the model for more efficient processing and performance on hardware accelerators.
This process requires a calibration dataset with inputs that are representative of the inputs that will be used on the specific model. 
The Converter also provides an interface that unburdens the user from transforming the dataset to the required AI-framework format.
The user only needs to provide the dataset in the \texttt{tf.data.Dataset} form.

\textbf{Composer:}
Next, the Composer is the second stage of the model-variant generation.
At this stage, user-provided input, (i.e. configuration files, interface implementation) and TF2AIF platform-specific code are combined with the consistent across all platforms and models Global Server Code (Fig. \ref{fig_tf2aif_agx_single_example}).

The platform- and AI-framework-specific components are the converted model from the previous stage (\circled{2}), the Base Server, and the Base Image.
The Base Server provides a class with specific functionality for the target AI-framework-platform combination. 
It includes methods that correspond to server initialization configurations and the model execution.
The standardization of the Base Server decouples the server functionality which depends on the AI-framework-platform intricacies from the functionality which is independent of the combination. 
This makes the design more modular and enhances the extensibility of TF2AIF by simplifying the process of introducing new AI-framework-platform combinations to TF2AIF.

To streamline the process of model-variant generation, TF2AIF includes a container template (Base Image) that sets the environment of the resulting server container.
While each AI-framework-platform combination needs to adhere to specific environment requirements, the focus is on minimizing these differences.
To achieve this, we select the same libraries and environmental variable configurations, where possible.
This eliminates any performance volatility introduced by variable library versions, etc. when experimenting with different model variants.
This feature is critical when TF2AIF is employed for hardware/model performance evaluation.

The second input of the Composer is the user-provided input (blue-shaded area in Fig. \ref{fig_tf2aif_agx_single_example}).
In more detail, the user can implement an interface related to the pre/post-processing of data, as well as provide configuration files for customization.
TF2AIF aiming to minimize the user intervention as much as possible, 
provides methods that implement functionality that changes per model such as the pre/post-processing functions.
In the evaluated model variants (Section \ref{sec:evaluation}), this interface was completed using only 100 lines of elementary Python scripting. 
Additionally, this interface is AI-framework-agnostic, thus it is reused across all AI-framework-platform combinations.
One more key feature of TF2AIF is the volume of customization options it provides.
Users can optionally define their custom specifications.
Those specifications include networking-related information, (e.g. IP address), model-related, (e.g. batch size), or AI-framework-related, (e.g. precision, parallelization factors).
TF2AIF is highly extensible, allowing users to add external files, libraries, and environmental variables.

\subsection{Features Outline}
\mycomment{
Goals:
\begin{itemize}
    \item Performant implementation (use of AI-frameworks)
    \item Ease of use: 1) framework-platform agnostic (interfaces that allow pre/post processing and quantization, existence of Base Images help) 2) Simplicity to implement interfaces (simple python scripting)
    \item Broad Hardware Support, especially edge (we implement ARM and AGX and use INT8 to facilitate)
    \item Extendibility: 1) Add more AI-framework-platform combinations (Our structure that decouples framework-platform specific parts from the rest of the functionality,  helps to extend) 2) extend created containers (files, libraries etc.)
    \item Reduce verification efforts
    \item Open Source
    \item Uniformity of different implementations (Highlight model differences)
\end{itemize}
}
\textit{Feature 1: Broad Hardware Support}. TF2AIF supports devices from all across the cloud-edge continuum, with a special emphasis on support for edge platforms. 
This broad hardware compatibility ensures that AI services can be deployed efficiently across a wide spectrum of devices, from high-end server accelerators to resource-constrained edge devices.
This broad hardware support addresses \textit{Objectives \#1} and \textit{\#3}, and distinguishes TF2AIF from existing solutions like Nvidia Triton and INFaaS, which exhibit limitations in edge hardware support.

\textit{Feature 2: Ease-of-use.} 
The design of TF2AIF prioritizes user-friendliness, being agnostic to specific AI-frameworks and platforms. It offers interfaces that simplify model pre/post-processing and quantization, allowing users to implement these features with basic Python scripting. The server base images abstract away the complexities of creating functional platform environments. This design significantly reduces the learning curve, thereby making TF2AIF accessible to a wider audience without necessitating deep understanding of underlying AI-frameworks or platforms. This feature is crucial in achieving \textit{Objective \#1} and sets TF2AIF apart from Nvidia Triton and INFaaS, which require comprehensive framework knowledge for similar implementations.

\textit{Feature 3: Performant Implementations.} Leveraging state-of-the-art AI-frameworks, TF2AIF attempts to maximize performance in AI inference tasks. By supporting advanced model optimization techniques such as quantization, TF2AIF optimizes processing efficiency and enhances performance on a variety of hardware accelerators. This commitment to high-performance implementations is fundamental to fulfilling \textit{Objectives \#1} and \textit{\#2} positioning TF2AIF as a tool that does not compromise on performance, even in the most demanding environments.

\textit{Feature 4: Extendibility.} 
The architecture of TF2AIF is thoughtfully constructed to facilitate easy integration of new AI-framework-platform combinations, thanks to its modular design. 
This is also highlighted by its open-source nature. 
The use of user-provided configuration files enhances its flexibility, allowing for the addition of external files, libraries, and environmental variables as necessary. 
Such extendibility fulfills \textit{Objectives \#2} and \textit{\#3}, by enabling developers to customize TF2AIF to their needs.

\textit{Feature 5: Accelerated Deployment.} 
TF2AIF markedly reduces the time required to transition from model development to deployment. 
By automating model conversion and container composition processes, TF2AIF enables rapid and efficient generation of production-ready AI services. 
This feature directly contributes to \textit{Objectives \#2} and \textit{\#4}, facilitating large-scale data analysis on model implementations and enabling ML-driven scheduling solutions.

\textit{Feature 6: Client Creation}.
A distinctive capability of TF2AIF is its automatic generation of example client containers, which are compatible with the server containers it produces. 
This feature not only facilitates the verification of AI inference services, but also their rapid deployment and integration into existing systems. 
The client container is created by a separate composer requiring minimal configuration from the user, thus aligning with \textit{Objective \#1}.

\section{Evaluation}
\label{sec:evaluation}

\subsection{Experimental Setup}

\newcolumntype{M}[1]{>{\centering\arraybackslash}m{#1}}

\begin{table}[t]
\centering
\caption{Experimental Setup}
\label{tab:devices}
\begin{center}
\begin{tabular}{|M{0.55cm}||M{1.7cm}|c|M{1.56cm}|M{2cm}|}
\hline
\textbf{}&\textbf{Architecture}&\textbf{CPU}&\textbf{Memory (GB)}&\textbf{Accelerator}\\
\hline
\hline
  \textbf{NE-1}&Intel\textsuperscript{\textregistered} Xeon\textsuperscript{\textregistered} Silver 4210 CPU @ 2.20GHz&16&16& Xilinx Alveo U280 (FPGA)\\
  \hline
  \textbf{NE-2}&Intel\textsuperscript{\textregistered} Xeon\textsuperscript{\textregistered} Gold 6138 CPU @ 2.00GHz &16&16&NVIDIA V100 (GPU)\\
  \hline
  \textbf{FE-1}&NVIDIA Carmel Arm\textsuperscript{\textregistered}v8.2 64-bit&8&32&512-core NVIDIA Volta (GPU)\\
  \hline
\end{tabular}
\label{tab:setup}
\end{center}
\end{table}

\begin{table}[htb]
    \centering
    \caption{Model Characteristics}
    \begin{tabular}{|l|c|c|c|}
        \hline
            \textbf{Model}            & \textbf{CNN Type} & \textbf{Size (MB)} & \textbf{Flops (GFLOPs)} \\
        \hline
        \hline
        LeNet   & Tiny  & 0.38   & 0.001 \\
        MobileNetV1    & Small  & 18.37  & 1.14  \\
        ResNet50         & Medium  & 102.78 & 7.73  \\
        InceptionV4    & Large  & 177.71 & 24.55 \\
        \hline
    \end{tabular}
    \label{tab:model_characteristics}
\end{table}

We evaluate the generated AI inference engines of the tool on a real-world experimental setup.
Specifically, we use a Kubernetes cluster that comprises the nodes described in Table  \ref{tab:setup}.
\mycomment{The near-edge (NE) nodes are two Virtual Machines (VMs), with 16 vCPUs and 16GB of RAM each.
The two VMs are connected via passthrough with a GPU, and an FPGA respectively.
The far-edge (FE) node is an NVIDIA Jetson Xavier AGX, which includes an ARM processor and an embedded GPU.
}
We utilized device plugins from NVIDIA\footnote{\url{https://github.com/NVIDIA/k8s-device-plugin}} and Xilinx\footnote{\url{https://github.com/Xilinx/FPGA\_as\_a\_Service}} for integrating their respective hardware into our setup. 
However, given that these vendors do not provide native support for ARM-based devices, we incorporated these devices into Kubernetes' unified framework by extending the Kube-API. 
This integration was essential to ensure seamless interaction and management of diverse hardware types within the Kubernetes environment.

We select the image classification use case and a variety of models described in Table \ref{tab:model_characteristics}.
We used TF2AIF to generate the deployment-ready AI inference services for every AI-framework-platform and model combination. For the GPU implementations, we used FP16 precision, exhibiting the potential of the AI-framework which allows us to take advantage of the device-specific hardware (e.g. FP16 Tensor Cores) \cite{Tensor-cores}. 
For the rest, we followed the precision choices that are depicted in Table \ref{table:inference_acceleration_precision}.
\mycomment{
In the scope of this experiment, we opt for INT8 precision for the FE node to adhere to the power constraints XXX-REF.
We use FP16 precision for the NVIDIA V100 variants, following the device specifications XXX-REF, and FP32 for the x86 CPU variants.
}

\subsection{Development Time}

\begin{figure}[t]
\centering
\includegraphics[width=0.99\columnwidth]{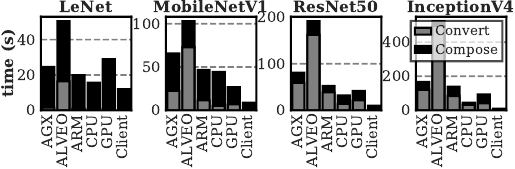}
\caption{AI service variants generation time; model conversion and final image creation.}
\label{fig:build}
\end{figure}

We run our model-variant generator on a physical machine equipped with a dual-socket Intel Xeon processor with 40 physical cores and 128 GB of RAM.
\mycomment{Additionally, \AT{na poume kai thn anagkh ths parousias ths GPU edw?}}
The system operates on Ubuntu 20.04.1 and utilizes Docker v20.10.21, inclusive of the buildx extension \cite{docker-buildx}. As detailed in Section \ref{sec:details}, the necessary procedures are conducted in parallel for each AI-framework-platform and for each model, optimizing the efficiency of the generation process.

Figure \ref{fig:build} illustrates the time required to generate model variants from the TensorFlow models listed in Table \ref{tab:model_characteristics}, across various platforms. 
The compose time is consistently between 20 and 40 seconds, while the conversion time greatly depends on the size of the provided model. 
As anticipated, the more lightweight models, such as LeNet and MobileNetV1, require less time for generation. 
It is noteworthy that the ALVEO version consistently demands the most time for preparation, which delay originates from the Vitis-AI conversion. 
In summary, our benchmarking tool efficiently produces 20 deployment-ready model variants, capable of serving AI inference services along with their respective clients, all within a span of just a few minutes.

\subsection{Benchmarking}

\begin{figure*}[t]
\centering
\includegraphics[width=0.85\textwidth]{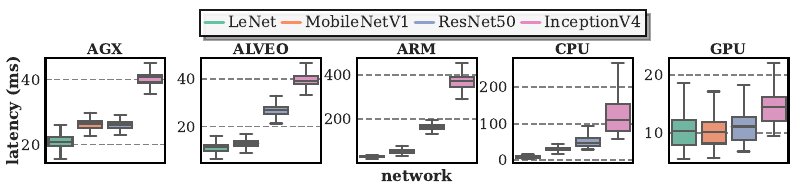}
\caption{Boxplot of the latency (ms) of each request for each AI-framework-platform model variant}
\label{fig:boxplot}
\end{figure*}

\begin{figure}[t]
\centering
\includegraphics[width=0.99\columnwidth]{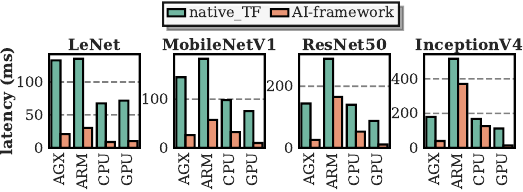}
\caption{Comparison of average latency (ms) between the selected AI frameworks, and the native TensorFlow implementations.}
\label{fig:latency_comparison}
\end{figure}

Performance evaluation of different model variations on different platforms is one of the use cases that TF2AIF excels at.
For the benchmarking, we use the generated clients, to request the different versions of AI service.
Each request includes one image.
To manage the deployment of the appropriate model variant upon request for a specific Artificial Intelligence Function (AIF), we have developed a backend system, which operates in conjunction with Kubernetes. 
This system, considering the available hardware, automatically determines the most suitable AI-framework-platform model variant for deployment.
A detailed analysis of this process is beyond this paper's scope.
We aim to explore the multi-objective node selection algorithm in future work, further refining this aspect of our implementation.

\mycomment{
We execute 1000 requests on each AI-framework-platform model variant and illustrate the boxplot of the execution latency in terms of milliseconds in Figure \ref{fig:boxplot}. 
We observe that for tiny or small models, like LeNet and MobileNetV1, the latency differences between platforms are diminished.
However, as models get more computationally intensive they take advantage of the increased capabilities of the platforms, which are reflected in the performance differences between platforms. 
Additionally, the least consistent AI-framework-platform in terms of latency appears to be the CPU \textcolor{blue}{(why?)}
}

For our performance benchmarking, we conducted 1000 requests for each AI-framework-platform model variant, focusing on the execution latency measured in milliseconds. 
The analysis, as depicted in Figure \ref{fig:boxplot}, reveals interesting patterns. We observed that smaller models, such as LeNet and MobileNetV1, show minimal latency variation across different platforms. 
This indicates a uniform efficiency in handling less complex tasks across varied hardware.
Conversely, with more complex models like ResNet50 and InceptionV4, the superior capabilities of certain platforms become evident. 
These platforms effectively leverage their advanced hardware to handle the increased computational demands, as reflected in their performance metrics.
Notably, the CPU AI-framework-platform demonstrated the highest latency variability, possibly due to system noise (e.g. context switching).

Additionally, to evaluate the advantages of using the AI-frameworks in TF2AIF over native TensorFlow implementations, we conducted a comparative experiment.
We replicated the server containers for each scenario in Table \ref{table:inference_acceleration_precision}, substituting the specialized AI-frameworks with native TensorFlow, except for the ALVEO AI-framework-platform due to the lack of FPGA support in TensorFlow. 
Similar to our primary evaluation, we executed 1000 requests on each setup.
The results, illustrated in Figure \ref{fig:latency_comparison}, show a significant performance improvement when using TF2AIF. 
The AGX, ARM, CPU, and GPU implementations achieved average speedups of 5.5x, 2.7x, 3.6x, and 7.6x, respectively, compared to their native TensorFlow counterparts. 
These findings underscore the substantial benefits of employing specialized AI-frameworks in our TF2AIF tool for accelerated AI inference.

\section{Conclusion \& Future Work}

In this paper we present TF2AIF, a tool that facilitates the development and deployment of AI models in the heterogeneous Cloud-Edge continuum.
TF2AIF automates the model conversion between a variety of AI frameworks while providing support for a variety of platforms.
This provides potential benefits for developers, especially when operating on Beyond 5G clusters, where the heterogeneity of devices is evident. 


Future research can exploit the automated generation of model variants to benchmark AI models and hardware combinations, deriving key insights. This data is crucial for developing scheduling mechanisms that select model variants based on optimization criteria. The ease and speed of generating performance data are vital in empowering AI/ML-driven schedulers for efficient decision-making, highlighting the importance of quick data availability in dynamic AI/ML applications.

\section*{Acknowledgment}
The research work was partially funded by the EU Horizon 2020 research and innovation programme, under project AI@EDGE, grant agreement No. 101015922.
Also, it was supported by the Hellenic Foundation for Research and Innovation (HFRI) under the 3rd Call for HFRI Ph.D. Fellowships (Fellowship Number: 5349).



%

\bibliographystyle{IEEEtran}
\bibliography{IEEEtran5/refs.bib}




\end{document}